\documentclass[11pt]{article}

\usepackage[margin=1in]{geometry}
\usepackage{amsmath, amssymb}
\usepackage{float}        
\usepackage{subcaption}
\usepackage{booktabs}     
\usepackage{hyperref}
\usepackage{authblk}      
\usepackage{natbib}       
\usepackage{graphicx}     

\title{Soliton-like Waves in a Two-Dimensional Recurrent Spiking Neural Network
       with Weighted Spike-Timing-Dependent Plasticity}

\author[1]{Christophe Meessen\thanks{Correspondence: \href{mailto:christophe@meessen.fr}{christophe@meessen.fr}}}
\affil[1]{Independent researcher, computational neuroscience, France}

\date{}

\begin{document}
\maketitle

\begin{abstract}
We construct a minimal but biologically plausible spiking neuron model
operating in discrete time, combining multiplicative spike-timing-dependent
plasticity (WSTDP), divisive normalization of synaptic integration,
homeostatic threshold adaptation, and a one-step refractory period.
We show that this normalization admits a biologically plausible dendritic
implementation in which each binary junction operates using only locally
available information.

Assembling excitatory-inhibitory pairs of such neurons into a
two-dimensional recurrent network and applying periodic localized
stimulation, we find that the network spontaneously gives rise to stable,
self-propagating wave packets with the properties of dissipative solitons:
they maintain a stable spatial profile, propagate at constant speed, and
annihilate upon frontal collision.
Their emergence requires a geometric asymmetry between excitatory and
inhibitory connection radii, and initial inhibitory synapses stronger than
excitatory ones.
WSTDP engraves the direction of propagation into the synaptic weight
profile, so that the network learns by itself to sustain propagation in
one direction while suppressing the reverse.
When two sources are active simultaneously, the resulting waves annihilate
upon collision, defining a semi-persistent boundary whose position encodes
the relative phase and frequency of the two sources.

These results provide a minimal computational framework for studying the
emergence of cortical traveling waves, activity zone delimitation, and
spatial memory from local plasticity rules alone.
\end{abstract}


\section{Introduction}

Traveling waves of neural activity are a ubiquitous phenomenon in the cortex.
They have been observed during development, sleep, and sensory processing,
propagating across the cortical surface at speeds ranging from a few to tens of
millimeters per second \citep{ermentrout2001traveling, muller2018cortical}.
Their functional role remains debated, but accumulating evidence suggests that
they modulate local excitability, coordinate timing across distant regions, and
may serve as a substrate for information propagation
\citep{muller2018cortical, davis2020spontaneous}.

The mechanisms by which such waves emerge in recurrent neural circuits are
not fully understood.
Neural field models have provided analytical insight into traveling waves,
solitary bumps, and oscillatory patterns
\citep{amari1977dynamics, bressloff2012spatiotemporal}, predicting that
a Mexican-hat connectivity profile --- short-range excitation balanced by
longer-range inhibition --- is a key condition for spatially structured activity.
However, these models abstract away individual spikes and synaptic plasticity.

Spike-timing-dependent plasticity (STDP) is a temporally asymmetric
Hebbian learning rule in which synaptic modification depends on the precise
relative timing of pre- and postsynaptic spikes
\citep{bi1998synaptic, markram1997regulation, debanne2023stdp},
endowing spiking networks with the ability to detect causal sequences and
self-organize receptive fields \citep{song2000competitive, masquelier2007unsupervised}.
A multiplicative weight dependence ensures stability without hard bounds
\citep{vanrossum2000stable, morrison2008phenomenological}.
Traveling waves were shown to build synaptic patterns via STDP in feedforward circuits
\citep{bennett2015refinement} and to strengthen preferred pathways in recurrent networks
\citep{butler2025traveling}.
However, to our knowledge, no study has examined whether a recurrent E/I network
equipped with STDP can autonomously sustain stable propagating wave packets ---
soliton-like structures --- emerging from a localized stimulation and
exhibiting annihilation upon collision.

Two additional neuronal mechanisms complement the core dynamics.
This normalization — which expresses the synaptic input as a fraction
of the maximum possible excitatory drive and is sometimes used in
spiking network models for its stabilizing properties — differs from
the divisive normalization of \citet{carandini2012normalization};
here the denominator is the sum of synaptic weights, a quasi-static
quantity that evolves only through WSTDP. We provide a biologically
plausible dendritic mechanism by which it can be implemented.
The firing threshold undergoes activity-dependent homeostatic adaptation
inspired by the structural plasticity of the axon initial segment (AIS)
\citep{kole2012signal, debanne2018ais}, providing an escape from stall
states in which a neuron that does not fire cannot update its weights.
Recurrent connectivity is the rule rather than the exception in cortical
circuits: the dominant input to a cortical neuron originates from neighboring
excitatory neurons \citep{douglas2004neuronal}, and understanding the
computational principles that emerge from such architectures remains
a fundamental open question in neuroscience.

Here we present a computational study of a two-dimensional recurrent network of
excitatory-inhibitory (E/I) neuron pairs with a weighted variant of STDP (WSTDP).
Starting from a spatially uniform initial connectivity, a repeated localized
stimulation gives rise to stable, self-propagating wave packets that share key
properties with dissipative solitons: they maintain a stable spatial profile,
propagate at constant speed, and annihilate upon frontal collision.
We characterize the conditions for their emergence, examine the role of the
excitatory-inhibitory balance, and analyze the dynamics of synaptic weight
convergence under WSTDP.
We further explore how stimulation parameters --- frequency, phase, and the
interaction between two simultaneous sources --- influence the resulting wave
patterns, including collision dynamics and the emergence of semi-persistent spatial
boundaries encoded in the synaptic weight landscape.
Our results suggest that recurrent spiking networks with purely local plasticity
rules can sustain a rich repertoire of spatiotemporal dynamics that may be
relevant for understanding wave propagation in cortical circuits.

\section{Materials and Methods}

\subsection{Network architecture}

We simulated a two-dimensional recurrent network of $200 \times 100$ neuron pairs,
each pair consisting of one excitatory (E) and one inhibitory (I) neuron.
Neurons are arranged on a regular rectangular grid, with one E/I pair per grid
point, analogous to the pixels of a digital image.
The network has hard boundaries: neurons at the edges are not connected to
neurons on the opposite side (no periodic boundary conditions).

Each neuron is connected to all neurons within a circular disk of a given radius
centered on its grid position.
Three types of synaptic connections are present, with their respective
connection radii:
\begin{itemize}
    \item E$\rightarrow$E (excitatory to excitatory): radius $r_{EE} = 9$,
          i.e.\ a disk of diameter 19 grid points;
    \item E$\rightarrow$I (excitatory to inhibitory): radius $r_{EI} = 5$;
    \item I$\rightarrow$E (inhibitory to excitatory): radius $r_{IE} = 5$.
\end{itemize}
No I$\rightarrow$I connections are included, consistent with the predominant
connectivity found in cortical circuits \citep{somogyi2005defined}.
Excitatory neurons include self-connections (E$\rightarrow$E autapses).
The behavior of these autapses under WSTDP --- in particular whether they
undergo overall LTD or LTP --- is characterized by the synaptic weight profile
as a function of relative position (Fig.~\ref{fig:weight_profile}).

The asymmetry between $r_{EE}$ and $r_{EI} = r_{IE}$ is a key structural
feature of the model: excitatory neurons project over a larger spatial range
than inhibitory neurons, creating a functional analog of a Mexican-hat
connectivity profile at the network level, without any explicit spatial
weighting of synaptic strengths.

All synaptic weights are initialized to a uniform value across the entire
network surface: $w_0^E = 0.5$ for excitatory synapses (E$\rightarrow$E and
E$\rightarrow$I) and $w_0^I = 10$ for inhibitory synapses (I$\rightarrow$E).
The strong initial inhibitory weight is a geometric necessity: because the
inhibitory disk (radius 5) contains fewer neurons than the excitatory disk
(radius 9), each individual inhibitory synapse must be stronger to provide
sufficient net inhibition and prevent runaway excitation.
The network thus starts as a spatially uniform ``blank'' state, with no
pre-existing spatial structure beyond the asymmetry of connection radii.

\subsection{Neuron model}

Simulation proceeds in discrete time steps.
At each time step $t$, the state of every neuron is updated synchronously.

\paragraph{Membrane potential and spike generation.}
For each neuron, the total synaptic input is computed as the weighted sum of
spikes received from all presynaptic neurons at time $t$:
\begin{equation}
    s(t) = \sum_{j \in \mathcal{P}} w_j \, x_j(t),
    \label{eq:input}
\end{equation}
where $\mathcal{P}$ is the set of presynaptic neurons, $w_j$ is the synaptic
weight of connection $j$, and $x_j(t) \in \{0, 1\}$ is the binary spike of
neuron $j$ at time $t$.
Spikes are binary: a neuron either fires ($x = 1$) or remains silent ($x = 0$)
at each time step.

To normalize the input with respect to the total excitatory drive, we compute
the activity ratio:
\begin{equation}
    a(t) = \frac{s(t)}{S_E},
    \label{eq:activity}
\end{equation}
where $S_E = \sum_{j \in \mathcal{P}_E} w_j$ is the sum of all excitatory
synaptic weights onto the neuron ($\mathcal{P}_E$ being the subset of
excitatory presynaptic neurons).
This normalization constrains the maximum value of $a(t)$ to $1$ --- reached
when all excitatory presynaptic neurons fire simultaneously --- while allowing
negative values when inhibitory input dominates, since inhibitory weights are
not normalized by $S_E$.

A neuron fires at time $t$ if $a(t) \geq \theta(t)$ and it did not fire at
time $t-1$ (refractory period of one time step).
After each time step, the membrane potential is reset to zero, consistent with
the discrete-time approximation.

\paragraph{Adaptive threshold.}
The firing threshold $\theta$ adjusts dynamically to maintain the neuron in a
functional operating range.
The threshold update rule is:
\begin{equation}
    \Delta\theta = \eta_\theta \bigl(\theta_{\min} - \theta
                  + x(t)\,(\theta_{\max} - \theta)\bigr),
    \label{eq:threshold}
\end{equation}
where $\eta_\theta$ is the threshold learning rate, $\theta_{\min} = 0.10$
and $\theta_{\max} = 0.60$ are the lower and upper bounds, and $x(t)$ is the
spike output of the neuron.
The update is applied only when $a(t) > a_{\min} = 0.05$, preventing
neurons with no input from drifting toward $\theta_{\min}$ and becoming
hyper-excitable.
On silence the threshold decreases toward $\theta_{\min}$ (recovering
excitability and avoiding WSTDP stall states); on spiking it increases
toward $\theta_{\max}$ (spike-frequency adaptation).
In steady-state the threshold converges to
$(\theta_{\min} + \theta_{\max})/2 = 0.35$.

\paragraph{Biological plausibility of the normalization.}
The normalization by $S_E$ (Eq.~\ref{eq:activity}) can be implemented
by a distributed dendritic algorithm: at each binary junction between
two subtrees with excitatory weight sums $s_1$ and $s_2$, each subtree
applies an attenuation factor $s_1/(s_1+s_2)$ and $s_2/(s_1+s_2)$
respectively, both in $[0,1]$ and therefore compatible with passive
attenuation.
Each junction relies solely on locally available information ---
a necessary condition for biological plausibility.
A consequence of this algorithm is that the membrane potential would
remain bounded by 1 throughout the dendritic tree and at the soma,
whereas in biological neurons the membrane potential increases by summation as
signals propagate toward the soma.
We suggest that this discrepancy may reflect the fact that biological neurons
do not perform exact arithmetic, and that the dendritic tree may implement only
an approximation of this normalization --- a compromise between computational
precision and biological constraints.

\subsection{WSTDP learning rule}

Synaptic weights are updated according to a weighted spike-timing-dependent
plasticity rule (WSTDP).
For each presynaptic neuron $j$ and postsynaptic neuron $i$, let
$\Delta t = t_{\text{post}} - t_{\text{pre}}$ be the difference between their
most recent spike times.

When a postsynaptic spike occurs, all presynaptic neurons that fired within
the past $\tau_{\max}$ time steps contribute a long-term potentiation (LTP)
update; conversely, when a presynaptic spike occurs, all postsynaptic neurons
that fired within the past $\tau_{\max}$ time steps contribute a long-term
depression (LTD) update.

The weight change for a spike pair of age $i = t - t_{\text{spike}} \in
\{0, 1, \ldots, \tau_{\max}\}$ time steps is:
\begin{align}
    \Delta w^{+} &= f(i) \,
                   \bigl(a_{\text{LTP}} - b_{\text{LTP}} \, w\bigr)
                   \quad \text{(LTP, pre before post)},
    \label{eq:ltp} \\
    \Delta w^{-} &= f(i) \,
                   \bigl(a_{\text{LTD}} - b_{\text{LTD}} \, w\bigr)
                   \quad \text{(LTD, post before pre)},
    \label{eq:ltd}
\end{align}
where $a_{\text{LTP}}$, $b_{\text{LTP}}$, $a_{\text{LTD}}$, $b_{\text{LTD}}$
are non-negative constants, and $f(i)$ is a discrete geometric (exponentially
decaying) temporal kernel:
\begin{equation}
    f(i) = \eta \cdot \epsilon^{\,i/(\tau_{\max}-1)},
    \quad i \in \{0, 1, \ldots, \tau_{\max}\},
    \label{eq:kernel}
\end{equation}
with $\eta$ the learning rate and $\epsilon$ the attenuation factor controlling
the relative weight of distant spike pairs.
By construction, $f(0) = \eta$ and $f(\tau_{\max}) = \eta\,\epsilon$, so the
kernel decays from $\eta$ to $\eta\,\epsilon$ over the full window.
The kernel is precomputed as a lookup table, making each weight update a single
multiply-add operation.
The same kernel is used for LTP and LTD updates; the two events are
distinguished solely by the sign convention and the parameters $a$ and $b$.
The same parameter values are used for all synapse types (E$\rightarrow$E,
E$\rightarrow$I, I$\rightarrow$E).

The multiplicative dependence on $w$ ensures that weights remain bounded
without hard clipping.
The weight update rule (Eqs.~\ref{eq:ltp}--\ref{eq:ltd}) is a generalized
multiplicative formulation derived from the experimental data of
\citet{bi1998synaptic}, also proposed independently by
\citet{standage2007trouble}, and no instabilities were observed in our simulations.
The weight converges to a fixed point $w^*$ determined by the balance between
LTP and LTD events.
Denoting by $F^+$ and $F^-$ the time-integrated rates of LTP and LTD events
respectively, the equilibrium weight satisfies:
\begin{equation}
    w^* = \frac{a_{\text{LTP}}\,F^+ - a_{\text{LTD}}\,F^-}
               {b_{\text{LTP}}\,F^+ + b_{\text{LTD}}\,F^-}.
    \label{eq:wstar}
\end{equation}
The natural bounds of the weight are $w_{\max} = a_{\text{LTP}}/b_{\text{LTP}}$
and $w_{\min} = a_{\text{LTD}}/b_{\text{LTD}}$.
Parameter values used in this study are listed in Table~\ref{tab:wstdp}.

\begin{table}[htbp]
\centering
\caption{WSTDP parameter values.}
\label{tab:wstdp}
\begin{tabular}{llr}
\toprule
Parameter & Description & Value \\
\midrule
$a_{\text{LTP}}$ & LTP amplitude          & $2.0$ \\
$b_{\text{LTP}}$ & LTP weight decay       & $2 \times 10^{-4}$ \\
$a_{\text{LTD}}$ & LTD amplitude          & $1 \times 10^{-3}$ \\
$b_{\text{LTD}}$ & LTD weight decay       & $1.0$ \\
$\eta$           & Kernel amplitude       & $5 \times 10^{-3}$ \\
$\epsilon$       & Kernel attenuation     & $0.1$ \\
$\tau_{\max}$    & Maximum spike age (steps) & $7$ \\
$\eta_\theta$    & Threshold learning rate   & $0.1$ \\
\midrule
$w_{\max} = a_{\text{LTP}}/b_{\text{LTP}}$ & Natural weight upper bound & $10^4$ \\
$w_{\min} = a_{\text{LTD}}/b_{\text{LTD}}$ & Natural weight lower bound & $10^{-3}$ \\
\bottomrule
\end{tabular}
\end{table}

The large ratio between $w_{\max}$ and the initial weights ($w_0^E = 0.5$,
$w_0^I = 10$) ensures that convergence to a value well below $w_{\max}$
is unambiguous.

The condition $\tau_{\max} \geq T_{\text{stim}}$, where $T_{\text{stim}}$ is
the inter-stimulation interval, is required for weight convergence: it
guarantees that LTD events occur with sufficient frequency to balance LTP,
preventing all weights from saturating at $w_{\max}$.

\subsection{Stimulation protocol}

At each stimulation event, all excitatory neurons within a circular disk of
radius $r_{\text{stim}} = 3$ grid points centered on a chosen location are
forced to fire, regardless of their membrane potential or threshold.
This models a strong localized external drive, as might arise from a thalamic
input or a local sensory stimulus.

Stimulation is applied repeatedly at a fixed period $T_{\text{stim}} = 7$ time
steps.
In single-source experiments, the stimulation site is fixed at the center of
the network $(width/2,\, height/2)$.
In collision experiments, two independent stimulation sources are placed
symmetrically at $(width/4,\, height/2)$ and $(3 \cdot width/4,\, height/2)$,
with potentially different phases and frequencies, allowing the study of
wave-wave interactions along the horizontal axis.

\subsection{Simulation implementation}

The simulation was implemented in Go, using 20 concurrent goroutines for
parallelization across the neuron grid.
The network comprises 20,000 excitatory and 20,000 inhibitory neurons
(40,000 total), connected by approximately 7.9 million synapses with the
reference radii $r_{EE} = 9$, $r_{EI} = r_{IE} = 5$.
Each neuron maintains a ring buffer storing the spike history of its
presynaptic inputs and its own output spikes over the WSTDP time window
$\tau_{\max}$.
On a workstation equipped with an AMD Ryzen AI~9 HX Pro~370 processor,
1,000 simulation steps complete in approximately 11~seconds under normal
propagation conditions, and up to 25~seconds in high-activity regimes
exhibiting fibrillation.

\section{Results}

\subsection{Emergence and propagation of soliton-like waves}

\begin{figure}[htbp]
    \centering
    \begin{tabular}{cc}
        \includegraphics[width=0.45\textwidth]{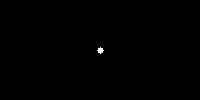} &
        \includegraphics[width=0.45\textwidth]{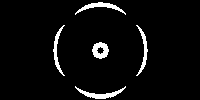} \\
        \small (A) $t = 0$, stimulation onset &
        \small (B) early propagation \\[1em]
        \includegraphics[width=0.45\textwidth]{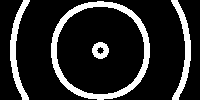} &
        \includegraphics[width=0.45\textwidth]{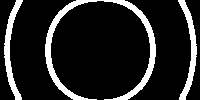} \\
        \small (C) advanced propagation &
        \small (D) $t > 270$, after stimulation offset \\
    \end{tabular}
    \caption{Snapshots of network activity during a single-source simulation
    (200$\times$100 network, stimulation site at center).
    White pixels indicate spiking excitatory neurons.
    \textbf{(A)} Initial stimulation at $t=0$.
    \textbf{(B--C)} Stable annular wave packets propagating outward during
    repeated stimulation.
    \textbf{(D)} After stimulation offset at $t=270$: inner rings disappear,
    confirming they are driven by the stimulus; the outermost wave packet
    continues to propagate autonomously.
    A supplementary video of the full simulation is available as
    Supplementary Movie~1.}
    \label{fig:snapshots}
\end{figure}

Starting from a spatially uniform connectivity and in the absence of any
pre-existing spatial structure, repeated localized stimulation rapidly gives
rise to stable, self-propagating wave packets.
Figure~\ref{fig:snapshots} shows four snapshots of network activity at
representative time steps during a single simulation of 300 time steps,
with stimulation applied at $T_{\text{stim}} = 7$ step intervals and
terminated at $t = 270$ (90\% of total duration).

At $t = 0$ (Fig.~\ref{fig:snapshots}A), the stimulation activates a small
disk of excitatory neurons at the network center, producing a localized
burst of activity.
Within the first few tens of time steps, this activity organizes into a
well-defined annular wave front that propagates radially outward from the
stimulation site (Fig.~\ref{fig:snapshots}B--C).
The wave packet maintains a stable spatial profile --- a bright excitatory
ring followed by an inhibitory trail --- as it travels across the network
surface at constant speed.
Multiple concentric rings are visible during active stimulation, each
corresponding to a successive stimulation event.

Critically, when stimulation is terminated at $t = 270$
(Fig.~\ref{fig:snapshots}D), the innermost rings disappear immediately,
confirming that they are sustained by the repeated drive and do not
propagate autonomously.
The outermost wave packet, however, continues to propagate without any
external input, demonstrating that it is self-sustaining.
This autonomous propagation is the hallmark of a soliton-like structure:
once established by the stimulation, the wave packet carries sufficient
synaptic drive to recruit the next layer of neurons through the
WSTDP-shaped connectivity, independently of the original source.

\subsection{Synaptic weight convergence under WSTDP}

\begin{figure}[htbp]
    \centering
    \includegraphics[width=0.85\textwidth]{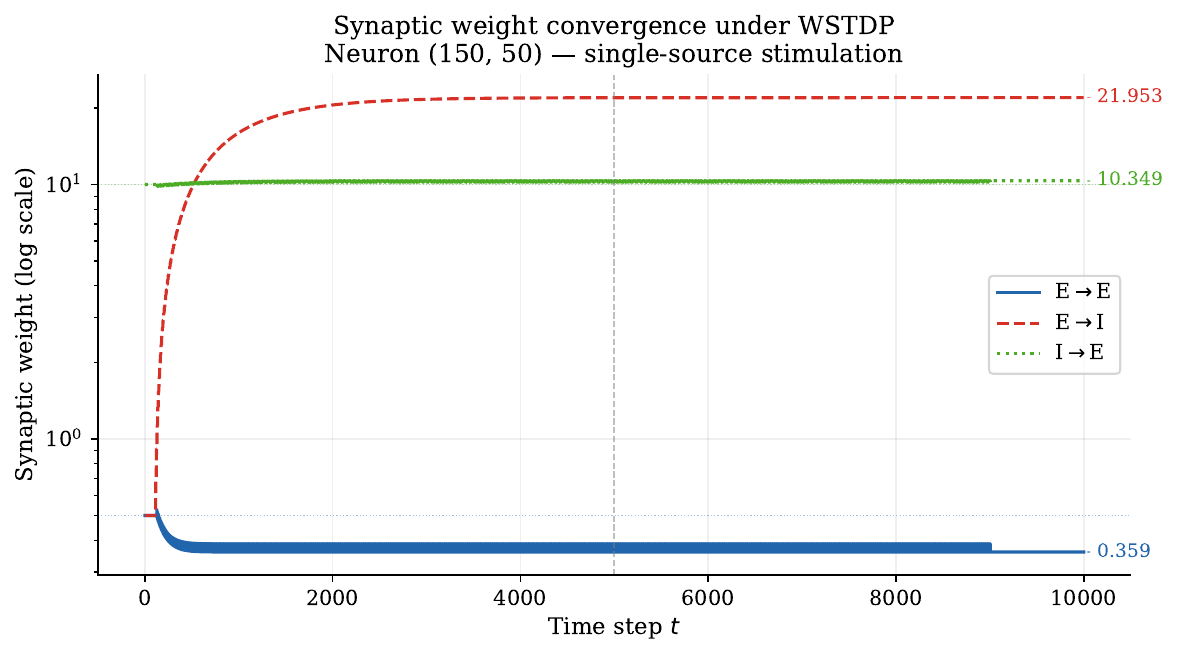}
    \caption{Evolution of synaptic weights over 10,000 time steps for neuron
    $(150, 50)$, monitoring outgoing synapses toward $(153, 50)$
    ($\Delta x = +3$, direction of soliton propagation).
    All three synapse types converge before $t = 5{,}000$ time steps.
    Dotted horizontal lines indicate initial weight values
    ($w_0^E = 0.5$ for E$\rightarrow$E and E$\rightarrow$I;
    $w_0^I = 10$ for I$\rightarrow$E).}
    \label{fig:weight_evolution}
\end{figure}

Figure~\ref{fig:weight_evolution} shows the evolution of three representative
synaptic weights over 10,000 time steps for neuron $(150, 50)$, monitoring
outgoing synapses toward $(153, 50)$ ($\Delta x = +3$, direction of propagation).
All three synapse types converge before $t = 5{,}000$.
The E$\rightarrow$E synapse converges to $w^* \approx 0.359$ (overall LTD);
the E$\rightarrow$I synapse to $w^* \approx 21.95$ (strong LTP in the direction
of propagation); the I$\rightarrow$E synapse to $w^* \approx 10.35$,
near its initial value.
These values are consistent with the fixed-point expression (Eq.~\ref{eq:wstar}).

\begin{figure}[htbp]
    \centering
    \includegraphics[width=0.85\textwidth]{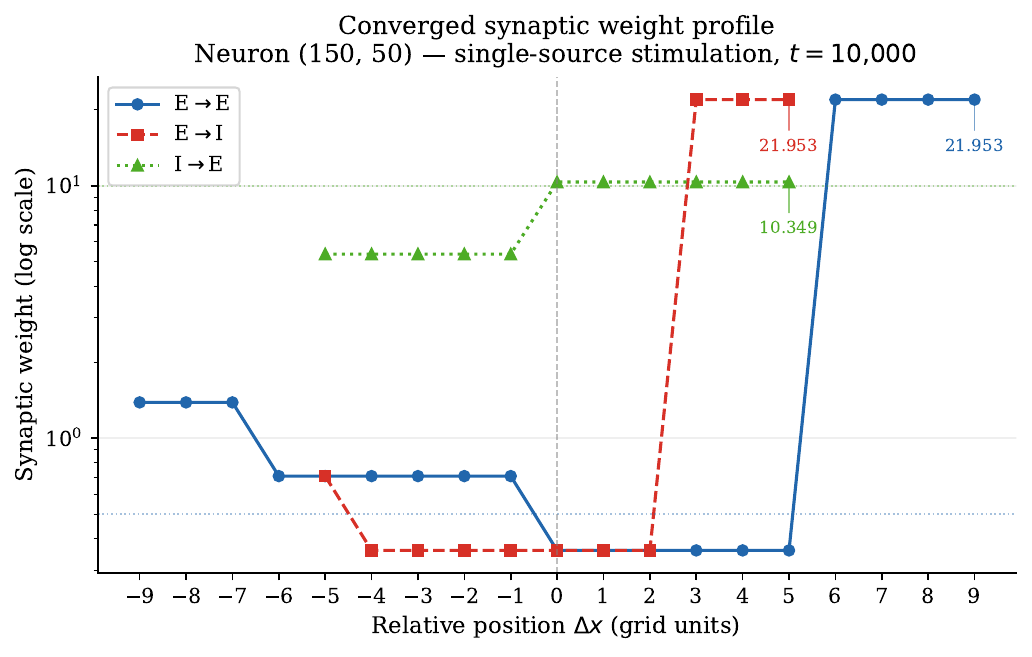}
    \caption{Converged synaptic weight profile along $y = 50$ for neuron
    $(150, 50)$, after $t = 10{,}000$ time steps.
    The strong left-right asymmetry reflects the causal structure of WSTDP:
    synapses in the direction of soliton propagation ($\Delta x > 0$)
    undergo overall LTP, while those in the opposite direction undergo overall LTD.
    $\Delta x = 0$ corresponds to the autapse (E$\rightarrow$E
    self-connection). Zero weights indicate absent connections and are
    not plotted.}
    \label{fig:weight_profile}
\end{figure}

Figure~\ref{fig:weight_profile} shows the converged spatial weight profile
along $y = 50$.
Three spatial zones are apparent for E$\rightarrow$E weights:
(i) the distal zone ($5 < |\Delta x| \leq 9$), beyond the reach of inhibitory
neurons, where weights converge to $\approx 1.39$ on the trailing side and
$\approx 10.35$ on the leading side;
(ii) the proximal zone ($|\Delta x| \leq 5$), where excitatory and inhibitory
inputs overlap, converging to lower values ($\approx 0.36$--$0.71$);
and (iii) the autapse ($\Delta x = 0$), converging to $\approx 0.36$.

The strong left-right asymmetry directly reflects the causal structure of
WSTDP: synapses in the direction of propagation are consistently activated
in a pre-before-post order (LTP), while those in the opposite direction
undergo overall LTD.
WSTDP thus engraves the direction of wave propagation into the synaptic
weight structure.

Weight convergence requires $\tau_{\max} \geq T_{\text{stim}}$: the WSTDP
window must be at least as large as the stimulation period to ensure
LTD events occur with sufficient frequency to balance LTP.

\subsection{Adaptive threshold profile}

\begin{figure}[htbp]
    \centering
    \includegraphics[width=0.85\textwidth]{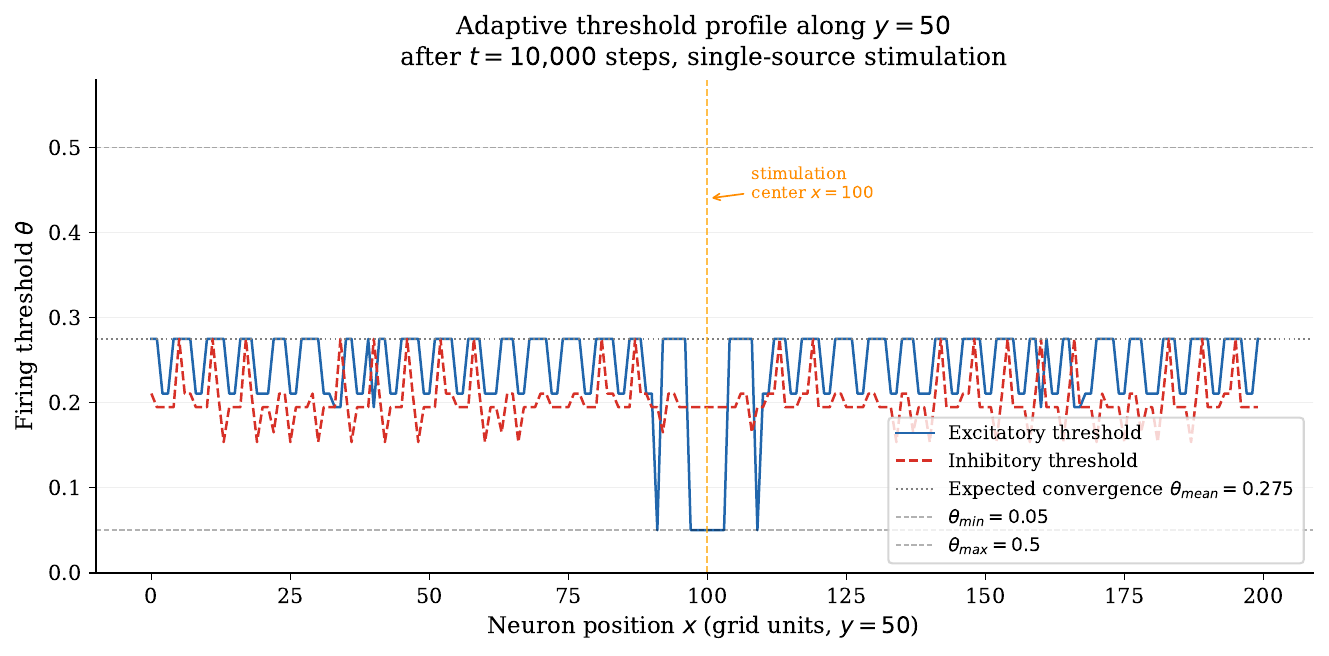}
    \caption{Adaptive firing threshold profile along $y = 50$ after
    $t = 10{,}000$ time steps.
    Most neurons converge near the expected steady-state value
    $\theta_{\text{mean}} = 0.275$ (dotted line).
    Neurons in the stimulation zone ($x \approx 91$--$109$) show thresholds
    at $\theta_{\min} = 0.05$, reflecting sustained direct activation.}
    \label{fig:threshold}
\end{figure}

Figure~\ref{fig:threshold} shows the converged firing thresholds along $y = 50$
after $t = 10{,}000$ steps.
Most neurons converge near $\theta_{\text{mean}} = 0.275$, confirming that the
adaptive threshold maintains neurons in their functional operating range.
Neurons in the stimulation zone ($x \approx 91$--$109$) converge to
$\theta_{\min} = 0.05$: repeatedly activated by direct stimulation, their
threshold rises on each spike then falls in the inter-stimulation interval,
with the net effect drifting to the minimum under sparse sustained activation.

\subsection{Collision dynamics and boundary persistence}

To investigate how soliton-like waves interact, we performed dual-source
experiments using two stimulation sites placed symmetrically at
$(width/4,\, height/2) = (50, 50)$ and $(3 \cdot width/4,\, height/2) = (150, 50)$.
Each source generates an outward-propagating wave packet; waves traveling
toward each other along the horizontal axis eventually collide.

\begin{figure}[htbp]
    \centering
    \begin{tabular}{cc}
        \includegraphics[width=0.45\textwidth]{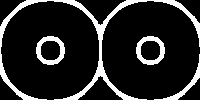} &
        \includegraphics[width=0.45\textwidth]{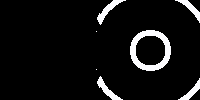} \\
        \small (A) frame 150, both sources active &
        \small (B) frame 312, left source silenced \\
    \end{tabular}
    \caption{Collision experiment: same frequency, same phase.
    \textbf{(A)} Two symmetric wave packets meet at the geometric midline
    $x = 100$ and annihilate.
    \textbf{(B)} With the left source silenced, the right soliton propagates
    freely but stops abruptly at the midline, revealing the persistent
    synaptic boundary encoded by WSTDP.}
    \label{fig:collision_same}
\end{figure}

\paragraph{Same frequency, same phase.}
When both sources are activated in phase at the same frequency
($T_{\text{stim}} = 7$), two symmetric annular wave fronts propagate inward
and meet precisely at the geometric midpoint $x = 100$
(Fig.~\ref{fig:collision_same}A, frame 150), where they annihilate.
The inhibitory trail of each soliton suppresses the excitatory front of the
other, preventing further propagation.

After stimulation of the left source is terminated
(Fig.~\ref{fig:collision_same}B, frame 312), the right soliton continues
to propagate but stops abruptly at the midline $x = 100$ even in the
absence of any competing wave.
The boundary is therefore not a transient collision point but a
\emph{persistent structural boundary} encoded in the synaptic weights by WSTDP:
the rule has strengthened synapses in the local propagation direction on
each side, while weakening cross-boundary synapses.
The normalization by $S_E$ then dilutes the contribution of the few
remaining forward synapses below threshold.

\begin{figure}[htbp]
    \centering
    \begin{tabular}{cc}
        \includegraphics[width=0.45\textwidth]{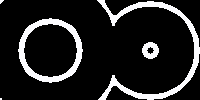} &
        \includegraphics[width=0.45\textwidth]{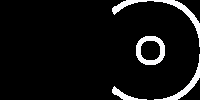} \\
        \small (A) frame 153, both sources active &
        \small (B) frame 315, left source silenced \\
    \end{tabular}
    \caption{Collision experiment: same frequency, phase offset of 3 steps
    ($\approx T_{\text{stim}}/2$).
    \textbf{(A)} The left source leads; its wave packets travel farther,
    shifting the annihilation boundary to the right of the midline.
    \textbf{(B)} With the left source silenced, the right soliton stops at
    the shifted boundary, confirming that boundary position encodes the
    relative phase of the two sources.}
    \label{fig:collision_diff}
\end{figure}

\paragraph{Same frequency, phase offset.}
When the right source is delayed by 3 time steps
($\approx T_{\text{stim}}/2$), the left source leads and its wave packets
travel farther before collision, shifting the annihilation boundary to the
right of the midline (Fig.~\ref{fig:collision_diff}A, frame 153).
When the left source is silenced (Fig.~\ref{fig:collision_diff}B, frame 315),
the right soliton stops at the shifted boundary, confirming that boundary
position encodes the relative phase of the two sources at the time of learning.

\subsection{Role of the excitatory-inhibitory balance}

The formation of soliton-like waves requires a precise balance between
excitation and inhibition.
This balance is determined not only by the synaptic weights but also by the
geometric asymmetry between connection radii: excitatory neurons project over
a larger area ($r_{EE} = 9$) than inhibitory neurons ($r_{IE} = 5$), so each
inhibitory synapse must individually be stronger to compensate for the smaller
number of inhibitory inputs.

\begin{figure}[htbp]
    \centering
    \includegraphics[width=0.6\textwidth]{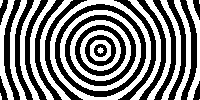}
    \caption{Network activity at frame 98 when the initial inhibitory weight
    is reduced to $w_0^I = 0.5$ (equal to the excitatory weight).
    Without sufficient inhibition to confine the excitatory wave front,
    the entire network oscillates synchronously, producing concentric rings
    that fill the whole surface.
    This epileptic-like regime contrasts sharply with the localized
    soliton-like waves observed with $w_0^I = 10$.
    The oscillation period reflects the stimulation period and the one-step
    refractory period of the neuron model.}
    \label{fig:fibrillation}
\end{figure}

To illustrate the consequence of disrupting this balance, we reduced the
initial inhibitory synaptic weight from $w_0^I = 10$ to $w_0^I = 0.5$,
matching the excitatory weight.
Figure~\ref{fig:fibrillation} shows the resulting network activity at
frame 98: instead of a localized propagating wave, the entire network
oscillates synchronously, with concentric rings of activity expanding from
the stimulation site and filling the whole surface.
This pattern resembles a global oscillation driven by the stimulation
periodicity and shaped by the one-step refractory period, which forces
alternating spiking and silence across the network.
No spatial confinement occurs, and no soliton forms.

This regime is analogous to an epileptic discharge in biological neural
circuits, where a failure of inhibitory control leads to runaway synchronous
excitation.
It confirms that the strong initial inhibitory weight $w_0^I = 10$ is not
an arbitrary parameter but a geometric necessity: it compensates for the
smaller spatial footprint of inhibitory connections relative to excitatory
ones, providing sufficient net inhibition to confine the excitatory wave
front and enable the formation of stable soliton-like structures.

\subsection{Effect of frequency difference between two sources}
\label{sec:freq}

To further characterize the interaction between competing wave sources, we
varied the stimulation frequency of the right source (stim$_2$, $x = 150$)
while keeping the left source (stim$_1$, $x = 50$) at the reference period
$T_1 = 7$.
Activity was recorded as a kymograph: the spiking state of all excitatory
neurons along $y = 50$ was sampled at each time step, producing a
space-time image in which the horizontal axis represents position $x$ and
the vertical axis represents time (top to bottom).
Diagonal streaks correspond to propagating wave fronts; their slope reflects
the propagation speed, and their vertical spacing reflects the stimulation
period.

\begin{figure}[H]
    \centering
    \begin{tabular}{cc}
        \includegraphics[width=0.40\textwidth]{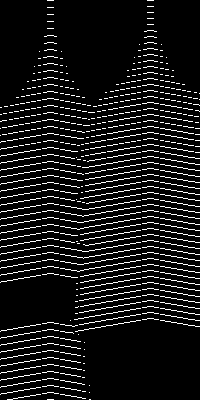} &
        \includegraphics[width=0.40\textwidth]{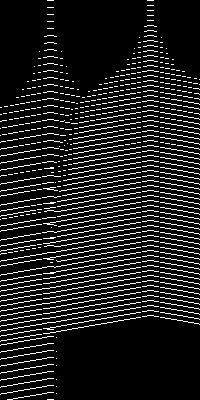} \\
        \small (A) $T_1=7$, $T_2=6$ & \small (B) $T_1=7$, $T_2=5$ \\
    \end{tabular}
    \caption{Kymographs of excitatory neuron activity along $y=50$ for
    dual-source experiments with different stimulation frequencies.
    Horizontal axis: position $x$ (0--200); vertical axis: time (top to
    bottom). White pixels indicate spiking neurons.
    Stim$_1$ (left source, $x=50$) has period $T_1=7$ in both panels;
    stim$_2$ (right source, $x=150$) has period $T_2=6$ \textbf{(A)} and
    $T_2=5$ \textbf{(B)}.
    \textbf{(A)} Small frequency difference: the boundary drifts slowly
    leftward, with both sources remaining active throughout.
    \textbf{(B)} Large frequency difference: the faster right source
    dominates, progressively erasing the left source territory; only a
    small triangular zone on the far left survives when stim$_1$ is
    silenced.}
    \label{fig:kymo_freq}
\end{figure}

\paragraph{Small frequency difference ($T_1 = 7$, $T_2 = 6$).}
Figure~\ref{fig:kymo_freq}A shows the kymograph for a right source with period
$T_2 = 6$.
The faster right source produces more closely spaced wave fronts and a curved
annihilation boundary that drifts slowly leftward, reflecting the progressive
advantage of the faster source.
The boundary has not stabilized by the end of the recording, suggesting that
with sufficient time the faster source would eventually submerge the slower one.

\paragraph{Large frequency difference ($T_1 = 7$, $T_2 = 5$).}
Figure~\ref{fig:kymo_freq}B shows the kymograph for $T_2 = 5$.
The faster source dominates decisively: the boundary moves rapidly leftward
and the territory of stim$_1$ shrinks continuously.
When stim$_1$ is silenced, only a small triangular zone on the far left
survives --- the region where solitons from both sources travel in the same
direction and cooperate rather than annihilate.
The synaptic weights built by stim$_1$ have been overwritten by the stronger
LTP flux from the faster source.

These results reveal a frequency-dependent territorial competition.
The boundary position reflects the relative LTP flux from each source,
proportional to its firing rate.
A sufficient frequency advantage produces complete territorial takeover,
while a small advantage yields a slowly drifting boundary --- suggesting
a threshold frequency ratio separating the two regimes.

\subsection{Effect of the inhibitory connection radii $r_{EI}$ and $r_{IE}$}

\begin{figure}[H]
    \centering
    \includegraphics[width=0.85\textwidth]{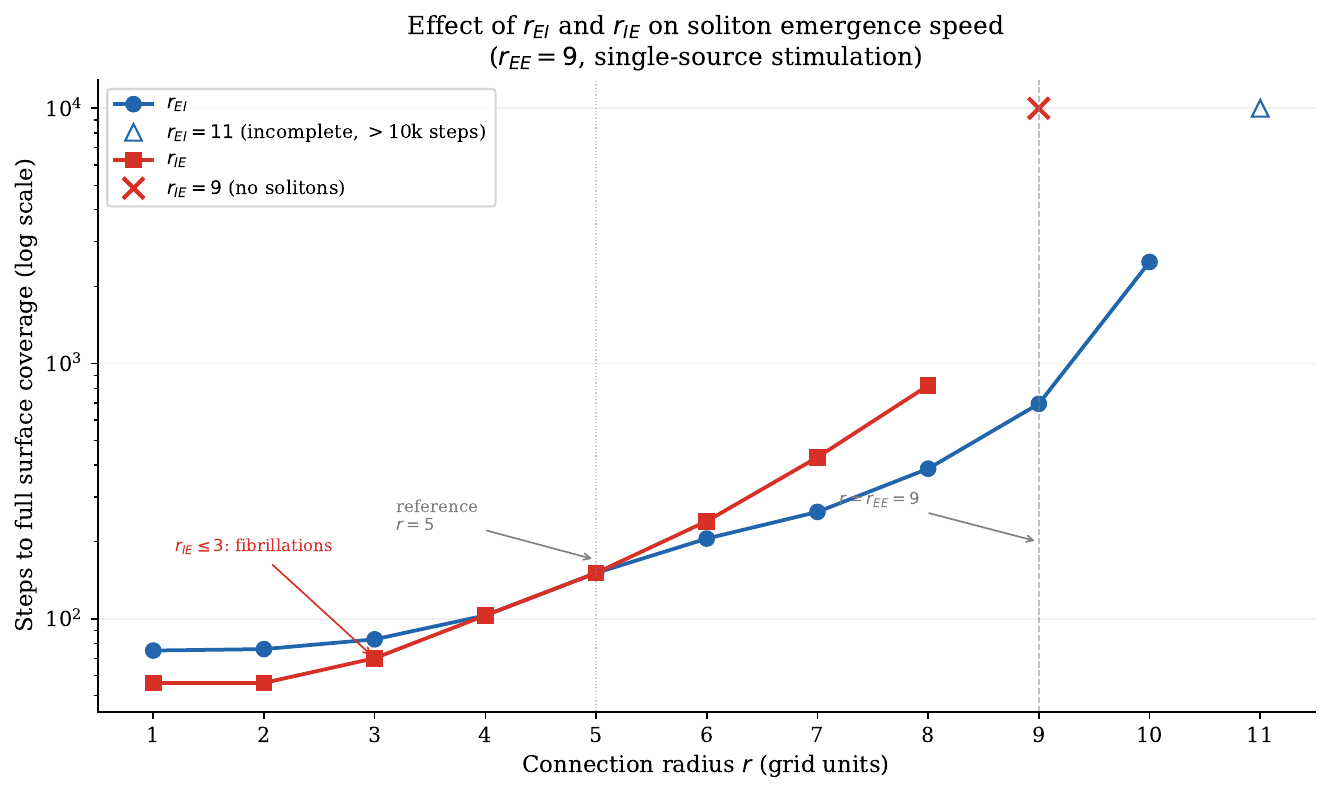}
    \caption{Effect of the inhibitory connection radii $r_{EI}$ (blue) and
    $r_{IE}$ (red) on soliton emergence speed ($r_{EE} = 9$, single-source
    stimulation).
    Both curves cross at the reference configuration $r = 5$ (dotted
    vertical line).
    For $r_{EI}$, solitons exist up to $r_{EI} = 10$ (narrower, $>$2500
    steps) but fail to cover the full surface at $r_{EI} = 11$ (open
    triangle).
    For $r_{IE}$, no solitons form at $r_{IE} = 9 = r_{EE}$ (cross).
    Fibrillations appear for $r_{IE} \leq 3$.
    All measurements are visual approximations.}
    \label{fig:rei_convergence}
\end{figure}

To characterize the role of the geometric asymmetry between excitatory and
inhibitory connection radii, we systematically varied each inhibitory
connection radius independently while keeping the excitatory radius fixed
at $r_{EE} = 9$.
For each value, we measured the number of time steps required for the
soliton to cover the full network surface.
Figure~\ref{fig:rei_convergence} shows the results for both $r_{EI}$ and
$r_{IE}$.

\paragraph{Effect of $r_{EI}$.}
For $r_{EI} \in [1, 9]$, the emergence time increases monotonically with
$r_{EI}$, ranging from approximately 75 steps ($r_{EI} = 1$--$2$) to
695 steps ($r_{EI} = 9 = r_{EE}$).
For $r_{EI} = 10 > r_{EE}$, solitons still emerge but require
approximately 2500 steps and are visibly narrower.
For $r_{EI} = 11$, the soliton fails to cover the full network surface
even after 10,000 steps, marking the boundary of the soliton existence
regime for this parameter.

\paragraph{Effect of $r_{IE}$.}
The $r_{IE}$ curve shows a similar monotonic trend for $r_{IE} \in [1, 8]$,
ranging from approximately 56 steps ($r_{IE} = 1$--$2$) to 820 steps
($r_{IE} = 8$).
However, two important differences distinguish it from the $r_{EI}$ curve.
First, for $r_{IE} \leq 3$, fibrillations appear after the initial soliton
formation: the reduced inhibitory footprint fails to fully suppress the
excitatory activity in the wake of the wave, leading to secondary
oscillations.
Second, for $r_{IE} = 9 = r_{EE}$, no solitons form at all --- a
qualitatively different outcome from the $r_{EI} = 9$ case where solitons
still emerge, albeit slowly.

\paragraph{Interpretation.}
Both curves cross at the reference configuration $r_{EI} = r_{IE} = 5$
(151 steps).
For $r < 5$, reducing $r_{IE}$ accelerates learning more than reducing
$r_{EI}$ by the same amount; for $r > 5$ the situation reverses, and
$r_{IE} = r_{EE}$ prevents soliton formation entirely while $r_{EI} = r_{EE}$
merely slows it.
This asymmetry suggests that $r_{IE}$ primarily controls inhibitory trail
confinement behind the wave front, while $r_{EI}$ controls lateral inhibition
ahead of it.

\subsection{Effect of the adaptive threshold parameters}

We investigated the influence of two threshold parameters on soliton
dynamics: the initial threshold value $\theta_{\text{init}}$ and the
maximum threshold $\theta_{\max}$, with the minimum threshold fixed at
$\theta_{\min} = 0.05$.
All previous simulations used $\theta_{\text{init}} = 0.15$ and
$\theta_{\max} = 0.50$, giving a steady-state mean threshold of
$(\theta_{\min} + \theta_{\max})/2 = 0.275$.

\begin{figure}[H]
    \centering
    \begin{tabular}{cc}
        \includegraphics[width=0.40\textwidth]{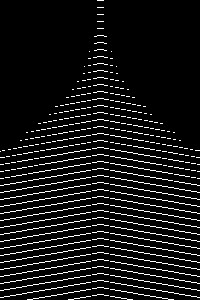} &
        \includegraphics[width=0.40\textwidth]{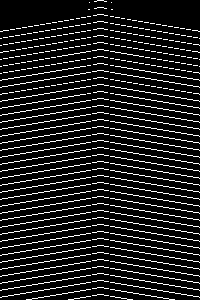} \\
        \small (A) $\theta_{\max}=0.50$, $\theta_{\text{init}}=0.15$ &
        \small (B) $\theta_{\max}=0.50$, $\theta_{\text{init}}=0.10$ \\[1em]
        \includegraphics[width=0.40\textwidth]{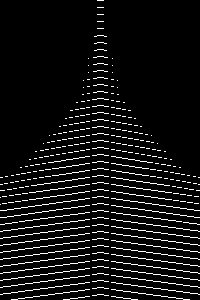} &
        \includegraphics[width=0.40\textwidth]{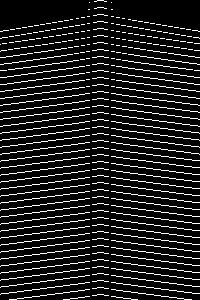} \\
        \small (C) $\theta_{\max}=0.10$, $\theta_{\text{init}}=0.15$ &
        \small (D) $\theta_{\max}=0.10$, $\theta_{\text{init}}=0.10$ \\
    \end{tabular}
    \caption{Kymographs of single-source stimulation for four combinations
    of $\theta_{\text{init}}$ and $\theta_{\max}$ ($\theta_{\min}=0.05$
    throughout).
    Horizontal axis: position $x$ (0--200); vertical axis: time
    (top to bottom).
    Reducing $\theta_{\text{init}}$ from 0.15 \textbf{(A, C)} to 0.10
    \textbf{(B, D)} markedly accelerates soliton emergence.
    Varying $\theta_{\max}$ has negligible effect on emergence speed.}
    \label{fig:kymo_thresh_single}
\end{figure}

\paragraph{Effect of the initial threshold on soliton emergence speed.}
Figure~\ref{fig:kymo_thresh_single} shows kymographs for four combinations
of $\theta_{\text{init}}$ and $\theta_{\max}$.
Reducing $\theta_{\text{init}}$ from $0.15$ to $0.10$ markedly accelerates
soliton emergence: neurons in the wave front are immediately excitable,
so WSTDP begins shaping the directional weight profile from the first
stimulation events rather than waiting for the dynamic threshold to descend.
By contrast, varying $\theta_{\max}$ between $0.10$ and $0.50$ has
negligible effect on emergence speed.

\begin{figure}[H]
    \centering
    \begin{tabular}{cc}
        \includegraphics[width=0.40\textwidth]{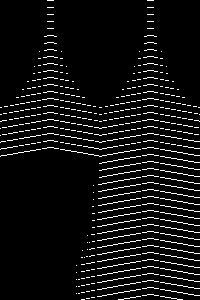} &
        \includegraphics[width=0.40\textwidth]{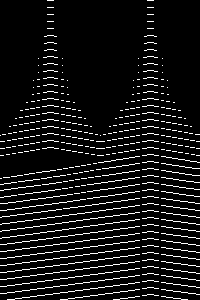} \\
        \small (A) $\theta_{\max}=0.50$ &
        \small (B) $\theta_{\max}=0.10$ \\
    \end{tabular}
    \caption{Kymographs of dual-source stimulation with stim$_1$ (left)
    silenced partway through, for $\theta_{\max} = 0.50$ \textbf{(A)}
    and $\theta_{\max} = 0.10$ \textbf{(B)} ($\theta_{\text{init}}=0.15$
    throughout).
    With high $\theta_{\max}$, the boundary is robust and erosion is slow
    \textbf{(A)}.
    With low $\theta_{\max}$, stim$_2$ rapidly invades the left territory
    after stim$_1$ is silenced \textbf{(B)}.
    In all cases solitons annihilate rather than cross; the mean threshold
    controls the rate of territorial erosion, not the annihilation itself.}
    \label{fig:kymo_thresh_dual}
\end{figure}

\paragraph{Effect of $\theta_{\max}$ on boundary persistence.}
Figure~\ref{fig:kymo_thresh_dual} shows dual-source kymographs with
stim$_1$ silenced partway through.
With $\theta_{\max} = 0.50$, the boundary is robust and stim$_2$ encroaches
only slowly into the left territory (Fig.~\ref{fig:kymo_thresh_dual}A).
With $\theta_{\max} = 0.10$, the boundary erodes rapidly: stim$_2$
progressively invades the left territory, the frontier receding leftward
over time (Fig.~\ref{fig:kymo_thresh_dual}B).
In both cases solitons annihilate rather than cross; the mean threshold
controls the rate of territorial erosion, not the annihilation itself.
A lower mean threshold ($\theta_{\text{mean}} = 0.075$ vs.\ $0.275$)
increases neuronal excitability, allowing the active source to recruit and
rewrite neurons in the opposing territory more easily.
The annihilation boundary is therefore semi-persistent: its robustness
is a continuous function of the mean firing threshold, which sets the
balance between stability and plasticity of learned territorial boundaries.

\section{Discussion}

\subsection*{A minimal model for emergent soliton-like waves}

A two-dimensional recurrent E/I network with WSTDP and spatially uniform
initial connectivity can spontaneously give rise to stable, self-propagating
wave packets upon repeated localized stimulation.
The model is deliberately minimal --- binary neurons, discrete time, uniform
initial weights --- suggesting that soliton-like emergence is a robust
property of recurrent E/I networks with local plasticity rather than an
artifact of specific parameter choices.

\subsection*{The causal role of WSTDP}

A central finding is that WSTDP is not merely permissive but causally
necessary for the observed phenomena.
The temporal window condition $\tau_{\max} \geq T_{\text{stim}}$ must be
satisfied for synaptic weights to converge: without sufficient LTD events,
all weights saturate at $w_{\max}$.
This constraint is specific to the highly regular stimulation used here;
with stochastic or naturalistic inputs, LTD events would occur continuously
and weight convergence would be expected regardless of window width.
The direction of wave propagation is engraved in the asymmetric weight profile
(Fig.~\ref{fig:weight_profile}): synapses in the direction of propagation
undergo overall LTP, while those in the opposite direction undergo overall LTD,
making the wave self-sustaining --- it carries its own ``track'' in the
synaptic weights.

The plasticity rule is a multiplicative, weight-dependent variant of STDP
that ensures bounded weights converging to a fixed point determined by
the ratio of integrated LTP and LTD event rates (Eq.~\ref{eq:wstar}).
The qualitative behavior --- soliton emergence, directional weight asymmetry,
persistent boundaries --- is governed by the ratio $F^+/F^-$ and is not
expected to depend on the specific parameter scale.

\subsection*{Directional stability and learned constraints}

A notable property of the network is that once a direction of propagation
has been established by WSTDP, propagation in the opposite direction becomes
impossible.
The asymmetric weight profile creates a strong directional bias: synapses
in the forward direction are potentiated, while those in the reverse
direction are depressed.
A wave attempting to propagate backward would find insufficient synaptic
drive to recruit the next layer of neurons, because the normalization by
$S_E$ dilutes the contribution of the few remaining forward synapses.

This directional constraint is functionally important: without it, each
excitatory wave front could trigger a return wave in the opposite direction,
leading to runaway oscillations analogous to fibrillation.
The WSTDP rule thus not only enables soliton propagation but actively
stabilizes it by suppressing unstable reverse modes --- a self-reinforcing
mechanism that emerges purely from local spike-timing interactions.

However, this directional constraint is not irreversible.
The frequency competition experiments (Section~\ref{sec:freq}) demonstrate that a
sufficiently persistent stimulation at higher frequency can progressively
overwrite the established weight profile and redirect propagation.
This suggests that the directional memory is robust against weak
perturbations but plastic under sustained drive --- a property reminiscent
of metaplasticity in biological synapses \citep{abraham2008metaplasticity}.

\subsection*{Dissipative solitons and persistent boundaries as spatial memory}

The wave packets described here share key properties with solitons ---
stable spatial profile, constant propagation speed, robustness to
perturbation --- but annihilate upon frontal collision rather than
passing through each other, which is characteristic of
\emph{dissipative solitons} found in non-equilibrium nonlinear systems
such as reaction-diffusion models and optical cavities.
The annihilation arises from the inhibitory trail that follows each
excitatory front, which suppresses the oncoming wave.

The collision experiments further reveal that the annihilation boundary is
a semi-persistent spatial structure encoded in the synaptic weights.
Once formed, it resists erasure even when one source is silenced, and its
position encodes the relative stimulation history: it reflects the phase
offset between in-phase sources and shifts progressively when sources have
different frequencies.
This phase- and frequency-dependent boundary formation suggests that
recurrent spiking networks with WSTDP implement a form of
\emph{spatial memory} --- encoding the relative timing and rate of past
inputs as persistent territorial boundaries in the synaptic weight
landscape --- potentially relevant for understanding how cortical circuits
segregate and maintain distinct activity domains.

\subsection*{Relation to existing models}

The present results complement and extend several lines of existing work.
Neural field models predict stationary localized bumps and traveling waves
in networks with Mexican-hat connectivity \citep{amari1977dynamics,
bressloff2012spatiotemporal}, but these models abstract away individual
spikes and plasticity.
Our model shows that similar spatiotemporal structures emerge in a discrete
spiking network without any pre-wired Mexican-hat profile: the effective
asymmetry arises solely from the geometric difference in connection radii
combined with WSTDP.
Bennett and Bair \citep{bennett2015refinement} showed that traveling
waves can shape synaptic connectivity via STDP in feedforward circuits; our
work demonstrates the converse --- that STDP in a recurrent circuit can
shape and sustain the waves themselves.
More recently, \citet{butler2025traveling} showed that STDP can strengthen
preferred propagation pathways for traveling waves in a two-dimensional
spiking network; however, their model uses a continuous-time simulator with
stochastic E/I assignment, making a direct comparison with our results
difficult.

\subsection*{Limitations}

The model incorporates several simplifications.
First, simulation proceeds in discrete time with synchronous updates,
approximating the continuous asynchronous dynamics of biological circuits;
whether soliton-like phenomena emerge under asynchronous propagation remains
an open question.
Second, each neuron emits at most one spike per step (no burst firing),
biologically motivated by the predominance of isolated cortical spikes
\citep{kruger1988multimicroelectrode}; burst firing could alter collision
quantitative details but is unlikely to enable soliton crossing, as the
directional weight profile persists regardless of the spike model.
The membrane potential is also reset at each step, eliminating leaky integration.
Third, stimulation is applied directly to excitatory neurons rather than via
synaptic inputs, minimizing free parameters; extending to synaptic inputs is
a natural next step.

\subsection*{Perspectives}

Two directions stand out for future investigation.
First, replacing direct stimulation with structured synaptic afferents would
allow study of how solitons are triggered, modulated, and potentially encode
input information.
Second, testing whether the phenomena reported here --- self-sustaining
propagation and collision annihilation --- persist under asynchronous spike
propagation and burst-capable neurons would constitute a strong test of their
generality.

\section{Conclusion}

We have presented a computational study of a two-dimensional recurrent
network of excitatory-inhibitory spiking neuron pairs with weighted
spike-timing-dependent plasticity (WSTDP).
Starting from a spatially uniform connectivity, repeated localized
stimulation gives rise to stable, self-propagating wave packets that
we characterize as dissipative soliton-like structures.
These wave packets maintain a stable spatial profile, propagate at constant
speed, and annihilate upon frontal collision.

The emergence and stability of these structures depend critically on the
geometric asymmetry between excitatory and inhibitory connection radii,
which provides the spatial structure necessary for wave confinement, and
on the WSTDP rule, which engraves the direction of propagation into the
synaptic weight profile.
Collision experiments reveal that the annihilation boundary is a
semi-persistent spatial structure that encodes information about the
relative phase and frequency of competing stimulation sources --- a form
of spatial memory implemented in the synaptic weight landscape, robust
against weak perturbations but plastic under sustained drive.

These results suggest that recurrent spiking networks with local plasticity
rules can sustain a rich repertoire of spatiotemporal dynamics, and
provide a minimal computational framework for studying wave propagation,
territorial competition, and spatial memory in cortical circuits.

\section*{Conflict of Interest Statement}

The author declares that the research was conducted in the absence of any
commercial or financial relationships that could be construed as a potential
conflict of interest.

\section*{Acknowledgments}

The author used Claude Sonnet 4.6 (Anthropic, claude.ai) as an AI writing
assistant for the drafting and editing of this manuscript, the generation
of matplotlib figures, and \LaTeX{} typesetting.
All scientific content, results, and conclusions are the sole
responsibility of the author.

\bibliographystyle{plainnat}
\bibliography{references}

\end{document}